\documentclass[a4paper,fleqn]{cas-dc}
\usepackage{amssymb}
\usepackage{amsmath}
\usepackage{subcaption}
\usepackage{bigstrut}
\usepackage{xcolor}
\usepackage{wrapfig}
\usepackage{xspace}
\usepackage{longtable}
\usepackage{multirow}
\usepackage{algorithm}
\usepackage{algorithmic}
\newcommand{\ours}{\texttt{MSLA}\xspace}

\usepackage[numbers]{natbib}

\def\tsc#1{\csdef{#1}{\textsc{\lowercase{#1}}\xspace}}
\tsc{WGM}
\tsc{QE}

\begin{document}
\let\WriteBookmarks\relax
\def\floatpagepagefraction{1}
\def\textpagefraction{.001}

\shorttitle{}    

\shortauthors{}  

\title [mode = title]{Enhancing Oracle Bone Inscription Recognition via Multi-Scale Layer Attention}  



%

\author[label1]{Chaowen Yan}
\author[label2]{Kaishen Wang}
\author[label1]{Yong Wang}
\author[label2]{Jianlong Xiong}
\author[label3]{Tao He}[
      orcid=0000-0001-9405-3979]
\cormark[1]
\ead{tao_he@scu.edu.cn}

\affiliation[label1]{organization={Academy of Chinese Traditional Culture, Sichuan Normal University},
            city={Chengdu},
            country={China}}            
\affiliation[label2]{organization={College of Computer Science, Sichuan University},
            city={Chengdu},
            country={China}}
\affiliation[label3]{organization={School of Artificial Intelligence, Sichuan University},
            city={Chengdu},
            country={China}}
            









\begin{abstract}
Oracle Bone Inscriptions (OBIs) recognition plays a crucial role in understanding ancient Chinese culture. However, accurately recognizing OBIs remains highly challenging due to their complex, irregular, and often degraded shapes. Traditional methods rely on expert knowledge and manual analysis, which are time-consuming and error-prone. Although deep learning has greatly advanced general image recognition, existing methods struggle to capture the fine-grained details and subtle variations inherent in OBIs, resulting in limited performance.  Even most recent and effective layer attention techniques are designed to capture fine-grained dependencies through enhanced inter-layer interactions, yet they still exhibit only marginal improvements in OBIs recognition.  To address these limitations, we propose \textbf{Multi-Scale Layer Attention (\ours)}, a novel paradigm that explicitly models both multi-scale and cross-layer feature interactions. By enriching the representation with fine-grained details across multiple spatial scales, \ours enables more accurate and robust OBIs recognition. Extensive experiments on large-scale OBIs datasets demonstrate that \ours consistently outperforms existing attention mechanisms while maintaining computational efficiency.
\end{abstract}



\begin{keywords}
Oracle Bone Inscriptions \sep Layer Attention Mechanism \sep Multi-Scale Recognition
\end{keywords}

\maketitle


\section{Introduction}

\begin{figure*}
    \centering
    \includegraphics[width=0.9\linewidth]{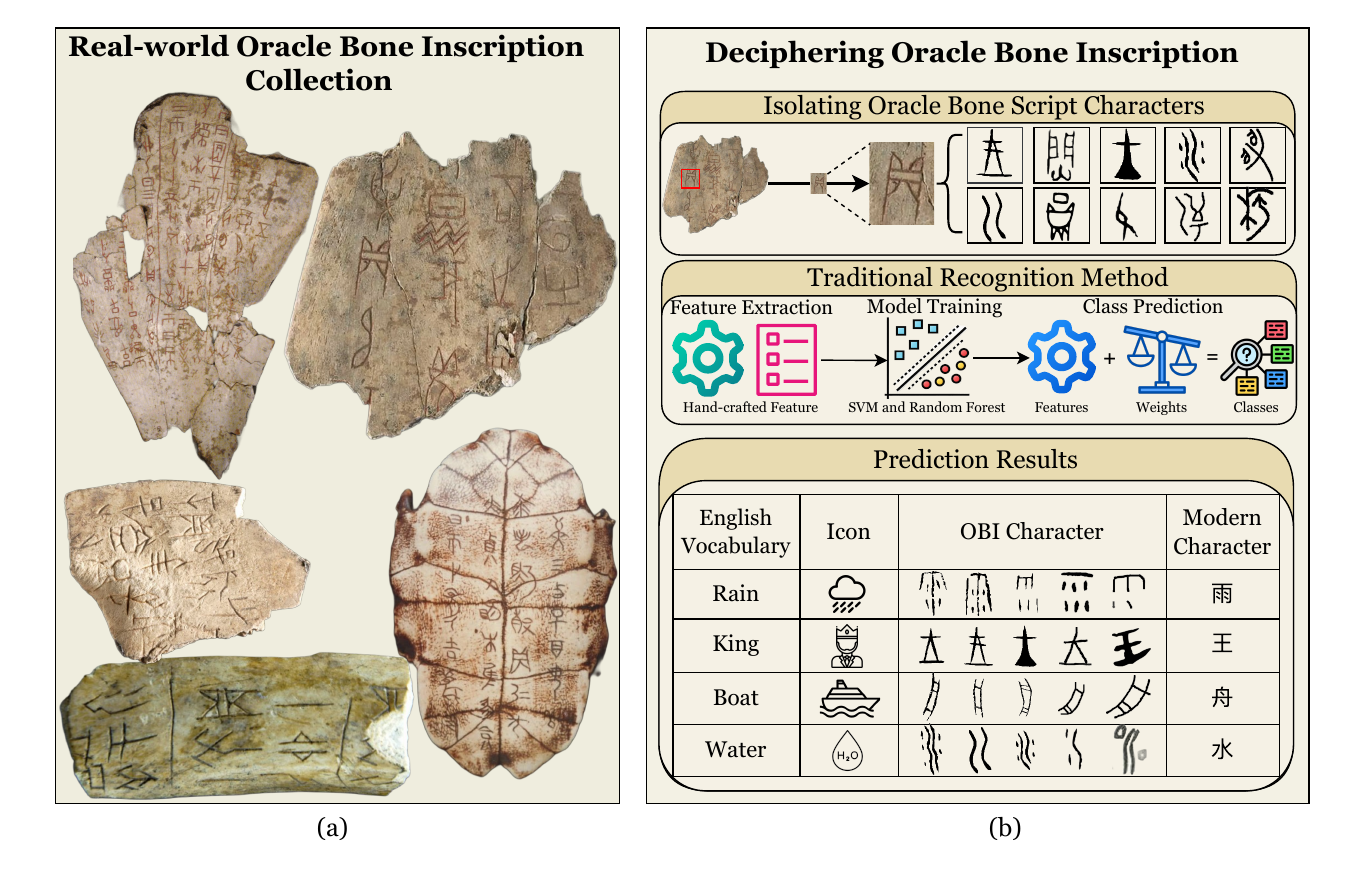}
    \caption{(a) Samples of real-world OBIs. (b) Deciphering OBI characters into modern Chinese characters via traditional recognition method.}
    \label{fig:OBI_example}
\end{figure*}

Oracle Bone Inscriptions (OBIs)~\citep{flad2008divination,kalinowski2009diviners,raphals2013divination}, shown in Figure~\ref{fig:OBI_example}(a), represent the earliest known systematized form of Chinese writing and carry profound historical and cultural significance, forming the foundation of modern Chinese characters. Accurate recognition of OBIs is essential not only for deciphering ancient historical records but also for advancing our understanding of the origins and evolution of Chinese civilization~\citep{keightley1979shang,boltz1986early}. However, the complex, irregular, and often fragmented structures of OBIs make automatic recognition particularly challenging.

Traditionally, the recognition of OBIs relied on expert knowledge and manual analysis~\citep{keightley1979shang,takashima2000towards,stough2011chinese,ojala2002multiresolution}, which is labor-intensive and prone to errors. With the advent of large-scale OBI datasets ~\citep{huang2019obc306, wang2024open, wang2024dataset,wang2022oracle}, deep learning techniques have been leveraged to improve recognition efficiency and accuracy. Architectures such as AlexNet \citep{krizhevsky2012imagenet}, ResNet \citep{he2016deep}, and ViT \citep{dosovitskiy2020image} have demonstrated remarkable performance in conventional image recognition tasks, offering potential for addressing the complex and diverse characteristics of OBIs through hierarchical feature learning.


To further enhance the model’s representational capacity and task performance, attention mechanisms~\citep {vaswani2017attention} have been incorporated into image recognition tasks. These approaches—such as channel attention~\citep{hu2018squeeze, woo2018cbam, wang2020eca}, spatial attention~\citep{wang2018non, carion2020end}, branch attention~\citep{srivastava2015training, li2019selective}, and temporal attention~\citep{xu2017jointly, chen2018video}—aim to make feature extraction more effective by selectively emphasizing the most informative channels, spatial regions, branches, or temporal components, thereby enhancing recognition performance. 

Although effective in general-purpose domains such as ImageNet~\citep{deng2009imagenet}, Cityscapes~\citep{cordts2016cityscapes}, and COCO~\citep{lin2014microsoft}, these methods exhibit limited performance when transferred to the specialized task of OBIs recognition. For instance, when applied to the large-scale OBIs dataset HUST-OBS~\citep{wang2024open} using ResNet-50 as the backbone, the aforementioned methods achieved comparable performance but introduced additional parameters and increased training costs. We attribute this phenomenon to the complex and diverse nature of OBIs—for example, variations in character morphology and degradation patterns—which makes them challenging to model using conventional attention mechanisms. This observation also indicates that existing methods struggle to capture the fine-grained details and subtle variations inherent in OBIs, thereby limiting their recognition performance. Motivated by this, we raise a new research question for OBIs recognition: \textit{Can we design a novel paradigm capable of capturing more fine-grained details to achieve accurate and robust OBIs recognition?}

While the most recent and effective attention paradigm in general visual recognition tasks is layer attention~\citep{fang2023cross, wang2024strengthening}, which dynamically incorporates preceding layers to capture global interactions by allowing each layer to attend to all its predecessors, thereby greatly enhancing inter-layer interaction effectiveness. This design naturally enables the model to capture richer and more fine-grained feature dependencies across layers, making it a promising direction for complex visual tasks. However, when applied to OBI recognition, we observe that despite its improvement over conventional attention mechanisms, layer attention still struggles to achieve excellent performance in OBIs recognition.

Upon further examination, we find that the representational capacity of existing layer attention mechanisms is inherently constrained by the limited number of tokens involved in cross-layer interaction. As shown in Table~\ref{tab:token_comparison}, unlike traditional attention mechanisms in Transformer~\citep{vaswani2017attention} that operate on hundreds of tokens, layer attention typically utilizes only a handful of tokens (e.g., fewer than 5 in ResNet backbones), resulting in insufficient diversity to fully encode cross-layer information. This token scarcity restricts the model’s ability to capture fine-grained dependencies, thereby weakening the potential advantages of layer attention in complex recognition tasks such as OBIs.

\begin{table}[htbp]
\centering
\caption{Comparison of approximate token counts and representational scope across different attention mechanisms.}
\label{tab:token_comparison}
    \begin{tabular}{lcl}
    \hline

    \hline
    Mechanism & Tokens & Representational Scope \\
    \hline
    Vision Transformer   & $\sim$196      & Global Patch Tokens \\
    Swin Transformer     & $\sim$49       & Window-based Tokens \\
    Non-local Attention  & $\sim H \times W$ & Full Spatial Positions \\
    Channel Attention    & $C$            & Channel Dimension \\
    Spatial Attention    & $\sim H \times W$ & Spatial Dimension \\
    Layer Attention      & $\sim 5$       & Multi-layer Features \\
    \hline

    \hline
    \end{tabular}
\end{table}

To address this, we propose Multi-Scale Layer Attention (\ours), a novel layer attention mechanism that explicitly models both multi-scale and cross-layer feature interactions. Unlike conventional layer attention, which operates on a limited number of tokens and thus restricts the diversity of cross-layer information, \ours enriches the token set by introducing multi-scale tokens derived from both global and local spatial representations.
Specifically, multi-scale feature extraction enables the model to capture fine-grained local details and holistic global context simultaneously, while a progressive accumulation strategy aggregates key and value representations from all preceding layers.
This design allows each layer to attend to a growing pool of multi-scale tokens, thereby achieving richer and more efficient cross-layer information exchange. Extensive experiments demonstrate 
that \ours enhances the model’s ability to capture subtle variations and complex structural dependencies in OBIs, leading to more accurate and robust recognition performance.

The contributions of this paper are summarized as follows:

\begin{enumerate}

\item We propose Multi-Scale Layer Attention (\ours), a novel attention paradigm designed for oracle bone inscription (OBI) recognition. \ours explicitly models both multi-scale and cross-layer feature interactions, addressing the inherent token scarcity in conventional layer attention mechanisms.

\item We introduce a multi-scale token enrichment and progressive accumulation strategy, which enables each layer to attend to a progressively expanding set of multi-scale tokens derived from both global and local representations. This design substantially enhances the model’s ability to capture fine-grained details and complex structural dependencies in OBIs.

\item We conduct extensive experiments on large-scale OBIs datasets, where \ours consistently outperforms state-of-the-art attention mechanisms while maintaining high computational efficiency. The results demonstrate the effectiveness, scalability, and general applicability of \ours to challenging fine-grained recognition tasks.
\end{enumerate}

\section{Related Work}
\subsection{Traditional OBIs Recognition}
Oracle Bone Inscriptions (OBIs) play a crucial role in the study of ancient Chinese civilization, as they record valuable information about early history, culture, and social activities. In particular, OBIs contain rich content related to astronomy~\citep{zhen1995astronomy}, ritual practices~\citep{flad2008divination}, and symbolic systems~\citep{zhang2021deciphering}, and have been recognized as an important medium for preserving and strengthening human collective memory~\citep{anzhu2020oracle}.

Traditional OBIs recognition methods primarily relied on conventional machine learning and image processing techniques. These approaches typically followed a two-stage pipeline, consisting of hand-crafted feature extraction followed by classical classifiers. Early studies often adopted template matching–based methods~\citep{meng2016recognition}, where target inscriptions were randomly selected from scanned rubbing images of oracle bone collections. The recognition process involved comparing the rubbing inscriptions with normalized character templates selected from an inscription database. These normalized templates were generated using character font software, resulting in smooth and clear characters with uniformly thick strokes and regular structures.

Subsequent works introduced more elaborate preprocessing and feature extraction strategies. For example, Meng et al.~\citep{meng2017recognition} applied Gaussian filtering and connected-component labeling to reduce noise, followed by affine transformation and thinning to extract character skeletons. Line feature points were then obtained using the Hough transform combined with clustering methods. Finally, recognition was performed by computing the minimum distance between the extracted line features of the input image and those of the template images.

More advanced traditional approaches focused on extracting robust hand-crafted features to improve recognition performance. These included structural and topological features, such as stroke skeletons and component layouts, as well as statistical descriptors. Widely used feature representations, including Histogram of Oriented Gradients (HOG)~\citep{dalal2005histograms}, Scale-Invariant Feature Transform (SIFT)~\citep{lowe2004distinctive}, and Local Binary Patterns (LBP)~\citep{ojala2002multiresolution}, were adapted to capture local texture and shape information from OBIs. Once feature vectors were extracted, classifiers such as Support Vector Machines (SVM) and K-Nearest Neighbors (KNN) were commonly employed to perform the final character classification.



\subsection{Deep Learning-based OBIs Recognition}
With the rapid development of deep learning, a wide range of deep learning–based methods have been applied to OBIs recognition. In particular, Convolutional Neural Networks (CNNs) have been extensively adopted due to their strong representation learning capability and outstanding performance in visual recognition tasks~\citep{liu2020oracle,mai2024oracle,fu2022improvement,gao2020distinguishing,meng2018recognition}. Specifically, Liu et al.~\citep{liu2020oracle} conducted comprehensive experiments to evaluate the performance of various CNN architectures on OBIs recognition, including AlexNet~\citep{krizhevsky2012imagenet}, VGG19~\citep{simonyan2014very}, SqueezeNet~\citep{iandola2016squeezenet}, ResNet~\citep{he2016deep}, and Inception-V3~\citep{szegedy2016rethinking}. Building upon this line of research, Mai et al.~\citep{mai2024oracle} further enhanced CNN-based models by employing parallel convolutional layers with different kernel sizes to better capture multi-scale features of OBIs.

Additionally, YOLO-based methods have also been widely applied to OBIs detection due to their high efficiency and strong real-time performance~\citep{xing2019oracle,xiao2025faa,li2024oracle,meng2024automatic,yan2024novel,fujikawa2023recognition,zhang2024rubbing}. 
For example, Fujikawa et al.~\citep{fujikawa2023recognition} integrated YOLO with MobileNet to achieve refined recognition of oracle bone characters. In their approach, YOLO is first employed to detect and recognize OBIs. However, since some characters may not be accurately detected in the initial detection stage, MobileNet is subsequently used to recognize the undetected OBIs after manually cropping them from the original images. 
Moreover, lightweight models, such as the improved YOLOv8n proposed in~\citep{zhang2024rubbing}, incorporate deformable convolutions and attention mechanisms to enhance feature extraction while maintaining computational efficiency, making them particularly suitable for large-scale OBIs analysis and cultural heritage applications.

Furthermore, recent efforts in OBIs recognition have focused on deep learning techniques to address challenges related to data scarcity and character complexity. The Structure-Texture Separation Network (STSN) \citep{wang2022unsupervised} effectively separates structural and textural features, enabling knowledge transfer from handprinted to scanned characters and enhancing recognition accuracy, even in the presence of noise. Additionally, unsupervised domain adaptation techniques such as pseudo-labeling and augmentation consistency \citep{wang2024oracle} improve model robustness against distortions and facilitate recognition of unlabeled scanned data. The Oracle Radical Extract and Recognition Framework (ORERF) \citep{lin2022radical} treats OBI characters as compositions of radicals, allowing for efficient radical detection via attention mechanisms.


\subsection{Attention Mechanisms in CNNs}
Recently, CNN-based models have been largely developed in many domains ~\citep{xu2026cnm,yu2026mobileode,he2024lightweight,cao2026enhancing,he2023cascade,he2024anchor}. 
Attention mechanisms~\citep{vaswani2017attention} have become an essential component in CNN-based models for enhancing feature representation by enabling networks to selectively emphasize informative features. In CNNs, attention mechanisms are generally categorized into channel attention, spatial attention, and layer attention, each focusing on different aspects of feature refinement. Channel attention~\citep{hu2018squeeze,qin2021fcanet,bastidas2019channel,wang2020eca} aims to recalibrate the importance of feature channels by modeling inter-channel dependencies, allowing the network to emphasize discriminative channels while suppressing less informative ones. For example, ECANet~\citep{wang2020eca} introduces a lightweight local cross-channel interaction mechanism without dimensionality reduction, achieving improved performance with minimal computational overhead. Spatial attention~\citep{wang2018non,zhu2019empirical,zhao2018psanet,carion2020end} focuses on identifying salient spatial regions within feature maps, enabling the model to capture long-range dependencies and contextual information by highlighting important spatial locations. By emphasizing meaningful regions, spatial attention improves object localization and contextual understanding in complex visual scenes. Layer attention~\citep{wang2024strengthening} assigns adaptive weights to features extracted from different network layers, facilitating effective fusion of multi-level semantic and spatial information. For instance, MRLA~\citep{fang2023cross} employs cross-layer interactions to dynamically aggregate features from multiple depths, leading to more robust representations.

Together, these attention mechanisms substantially enhance the representational capacity of CNNs and have demonstrated effectiveness across a wide range of computer vision tasks.



\begin{figure*}
    \centering
    \includegraphics[width=0.9\linewidth]{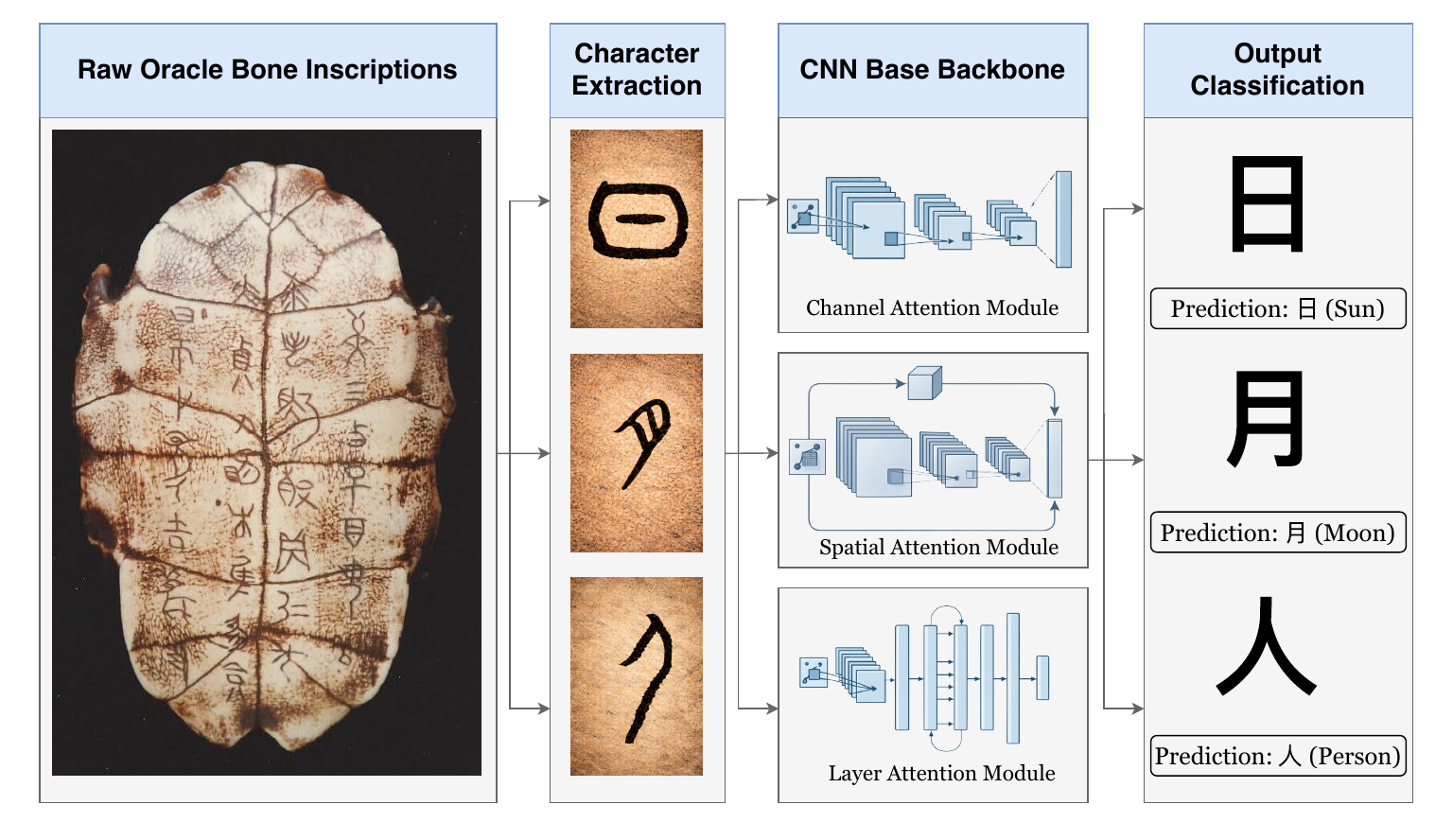}
    \caption{Deciphering oracle bone script inscriptions via Convolutional Neural Networks (CNNs).}
    \label{fig:deciphering_via_CNN}
\end{figure*}

\section{Method}
In this section, we first elaborate on existing layer interaction methods as preliminary concepts, followed by an architecture-level comparison. Building on the identified limitations, we then introduce the details of our proposed \ours.

\subsection{Problem Formulation}
Given an image of an OBI character, our objective is to identify its corresponding modern Chinese character. As illustrated in Figure~\ref{fig:OBI_example}(b), we present the traditional recognition methods and several examples of ancient OBIs alongside their modern counterparts. For instance, the first row of the prediction results corresponds to the ancient form of the Chinese character ``rain'', while the second row represents the character ``king''.

Formally, let the input OBI image be denoted as $x \in \mathcal{X}$. The goal is to learn a mapping function $f: \mathcal{X} \rightarrow \mathcal{Y}$, where $\mathcal{Y}$ denotes the set of modern Chinese character categories. Therefore, the task can be naturally formulated as a multi-class image classification problem, where the model predicts the correct modern character label $y \in \mathcal{Y}$ for a given input image $x$.

\paragraph{Traditional Methods.}
Under traditional paradigms, the OBI recognition task is typically decomposed into two stages: feature extraction and classification.

Given an input image $x$, a hand-crafted feature extractor $\phi(\cdot)$ is first applied to obtain a feature representation:
\begin{equation}
    z = \phi(x), \quad z \in \mathbb{R}^d.
\end{equation}
The extracted feature $z$ is then fed into a classifier $g(\cdot)$, such as a Support Vector Machine (SVM), to predict the corresponding modern character label:
\begin{equation}
    y = g(z) = g(\phi(x)), \quad y \in \mathcal{Y}.
\end{equation}
In earlier pattern-matching approaches, the mapping can be formulated as:
\begin{equation}
    y = \arg\min_{y' \in \mathcal{Y}} D\big(x, T_{y'}\big),
\end{equation}
where $T_{y'}$ denotes a template associated with class $y'$ and $D(\cdot,\cdot)$ is a predefined similarity or distance metric.

\subsection{Preliminary}

We formulate existing layer interaction methods into two patterns: layer aggregation and layer attention.

\paragraph{Layer Aggregation} 
Layer aggregation enhances inter-layer interaction through various novel modules applied along the network depth, which can be abstracted into a general paradigm. For brevity, we omit the skip connections in most neural network architectures. Given a feature $x_t$, denoting the output of the $t$-th layer, layer aggregation typically introduces a hidden representation $h_t$ to accumulate information across the depth of the network. This hidden state is iteratively updated by a lightweight transformation module $D_t$, parameterized by $\theta_t$, which integrates the current feature $x_t$ with the previous hidden state $h_t$, while simultaneously producing a refined version of $x_t$. The updated feature $x_t'$ is then passed into the $(t+1)$-th network block $F_{t+1}$. Formally, the procedure is defined as:
\begin{equation} \label{equ:layer_aggregation}
\left\{
\begin{aligned}
& x_t', h_{t+1} = D_t(x_t, h_t; \theta_t),  \\
& x_{t+1} = F_{t+1}(x_t').
\end{aligned}
\right.
\end{equation}

\begin{figure*}
    \centering
    \includegraphics[width=0.99\linewidth]{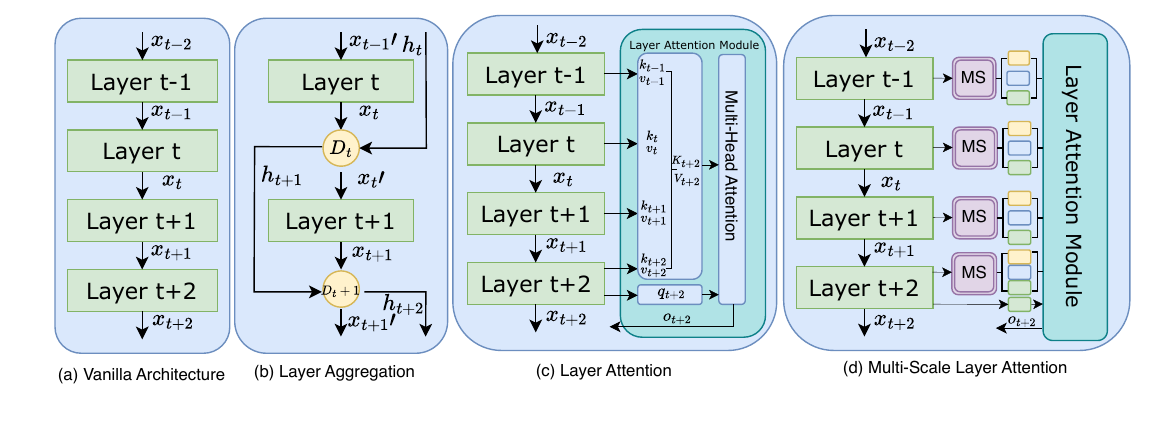}
    \caption{Architecture-level comparison among vanilla, layer aggregation, and layer attention. Skip connections are omitted for clarity.}
    \label{fig:architecture_comparison}
\end{figure*}

The architecture-level comparison between a vanilla neural network and layer aggregation is illustrated in Figure~\ref{fig:architecture_comparison} (a) and (b). Within this generalized formulation, representative methods such as DIANet \citep{huang2020dianet}, RLANet \citep{zhao2021recurrence}, and BA-Net \citep{zhao2022ba} can be naturally mapped. For example, in DIANet, the transformation module \( D \) is instantiated as a modified LSTM block with parameters shared across the network depth. In contrast, RLANet implements \( D \) as a lightweight Recurrent Layer Attention (RLA) block constructed from convolutional layers.

Although effective in certain scenarios, these methods are fundamentally constrained in two ways. First, their strictly forward-only design prevents each feature $x_t$ from accessing information from earlier layers, limiting the reuse of historical representations. Second, their localized interaction restricts $x_t$ to communicate only with its corresponding intermediate representation $h_t$, thereby narrowing the scope of cross-layer information exchange. Together, these limitations hinder the model’s ability to capture fine-grained details and long-range dependencies across the network depth.

\paragraph{Layer Attention} 
To overcome the limitations of forward-only propagation and localized interaction, layer attention has been introduced. 
In this paradigm, each layer is treated as a token, and an attention mechanism is applied to enable global interactions across all layers, thereby facilitating comprehensive cross-layer information exchange. 

The architecture is illustrated in Figure~\ref{fig:architecture_comparison} (c). Given the features \( x_i \) for \( i = 1, \dots, t \), which include all preceding layer features as well as the current layer feature \( x_t \), the query \( q_t \), key \( K_t \), and value \( V_t \) are derived as follows:  
\begin{equation} \label{equ:qkv_derivation}
\left\{
\begin{aligned}
q_t &= f_t^q(x_t), \\
K_t &= [f_1^k(x_1), \dots, f_t^k(x_t)], \\
V_t &= [f_1^v(x_1), \dots, f_t^v(x_t)].
\end{aligned}
\right.
\end{equation}

Here, \( f_i^k(\cdot) \) and \( f_i^v(\cdot) \) are functions used to compute the key vectors \( k_i \) and value vectors \( v_i \) for each feature \( x_i \). The key and value vectors are then concatenated to form the aggregated key representation \( K_t \) and value representation \( V_t \), respectively.

The attention output \( o_t \) is computed as:  
\begin{equation} \label{equ:2}
o_t = \text{softmax}\left(\frac{q_t K_t^\top}{\sqrt{d}}\right) V_t,
\end{equation}
where \( d \) is the dimensionality of the key vectors. The resulting output \( o_t \) refines the feature \( x_t \) at the \( t \)-th layer by integrating information from all preceding layers. By iteratively applying layer attention throughout the network depth, models can more broadly and effectively capture both fine-grained details and long-range dependencies across layers. Representative methods such as MRLA~\citep{fang2023cross}, DLA~\citep{wang2024strengthening}, and ELA~\citep{li2025enhancing} follow this formulation.

\subsection{Motivation}
Though layer attention alleviates the limitations of layer aggregation, such as the forward-only propagation and localized interactions that restrict cross-layer information exchange, it still suffers from drawbacks that undermine the efficiency of information interaction. 

In conventional attention mechanisms applied to domains like Natural Language Processing (NLP) or Computer Vision (CV), the number of tokens is typically large, enabling attention weights to encode rich and informative dependencies. As shown in Table~\ref{tab:token_comparison}, the standard Vision Transformer (ViT) employs 196 tokens. By contrast, layer attention modules operate with a much smaller token set. For instance, in a ResNet-20 backbone, a single stage provides at most 2 tokens for querying, while ResNet-56 increases this number only to 6. Such a limited token space constrains the diversity and richness of information that attention can capture, thereby reducing the effectiveness of cross-layer interactions.




Motivated by this observation, we propose to enrich layer attention by expanding the number of tokens available for interaction. However, merely increasing token quantity is insufficient, while additional tokens must faithfully represent the information embedded in earlier layers. To this end, we generate supplementary tokens by capturing features at multiple scales, ensuring that each new token encodes a distinct yet complementary perspective of the original representations. This design not only alleviates the limitation of insufficient tokens in layer attention but also introduces fine-grained information, thereby enabling the model to more effectively leverage cross-layer dependencies.

\subsection{The Proposed \ours}
To overcome the above limitations, we propose \ours, a novel layer attention mechanism that enriches the token set with multi-scale features. Convolutional operations with different receptive fields are used to extract multi-scale information, which expands the key and value representations. The current layer then attends to this enlarged feature space, and the resulting attention output refines its representation, enabling more effective cross-layer information exchange. The overall architectures are provided in Algorithm~\ref{alg:msla}.

\paragraph{Multi-Scale Token Extraction}
 
Unlike conventional attention mechanisms that operate on single-scale tokens, our approach enriches the attention process by introducing tokens at multiple scales from the input feature map \( x \in \mathbb{R}^{C \times H \times W} \), where $C$, $H$, $W$ represent channel, height, and width, respectively. This design ensures that both global contextual information and localized fine-grained patterns are preserved, thereby providing a more comprehensive token set for subsequent layer attention.  

To obtain a global representation, we first apply global average pooling to the feature map:  
\begin{equation}
x_{\text{global}} = \frac{1}{H \times W} \sum_{i=1}^{H} \sum_{j=1}^{W} x(:, i, j),
\end{equation}
where \( x_{\text{global}} \in \mathbb{R}^{1 \times C} \) serves as a holistic descriptor of the entire spatial domain.  

In parallel, to capture fine-grained spatial structures, the feature map is partitioned into a grid of \( P \times P \) non-overlapping patches, where \(P\) is a predefined patch division parameter. Each patch thus covers a region of size approximately \(\frac{H}{P} \times \frac{W}{P}\). For each patch, we apply adaptive average pooling to reduce it into a compact representation, resulting in  
\begin{equation}
x_{\text{local}} = \text{Pool}_{\text{adaptive}}(x, P), \quad 
x_{\text{local}} \in \mathbb{R}^{N_p \times C},
\end{equation}
where \(N_p = P \times P\) denotes the number of local tokens. Each local token encodes the average information of a specific spatial region, thereby complementing the global token with fine-grained details.

\begin{algorithm}[htbp]
\caption{Multi-Scale Layer Attention (\ours)}
\label{alg:msla}
\begin{algorithmic}[1]
\REQUIRE Feature maps $\{x_t\}_{t=1}^{T}$, patch division $P$
\ENSURE Refined feature maps $\{x_{t+1}\}_{t=1}^{T}$

\STATE Initialize $K_0 \leftarrow \emptyset$, $V_0 \leftarrow \emptyset$

\FOR{$t = 1$ to $T$}

    \STATE \textbf{// Multi-Scale Token Extraction}
    \STATE $x_{\text{global}} \leftarrow \text{GAP}(x_t)$
    \STATE $x_{\text{local}} \leftarrow \text{AdaptivePool}(x_t, P)$
    \STATE $X_{\text{multi},t} \leftarrow [x_{\text{global}}, x_{\text{local}}]$

    \STATE \textbf{// QKV Projection}
    \STATE $q_t \leftarrow W_q(X_{\text{multi},t})$
    \STATE $k_t \leftarrow W_k(X_{\text{multi},t})$
    \STATE $v_t \leftarrow \text{MultiScalePool}(W_v(x_t))$

    \STATE \textbf{// Progressive Key-Value Accumulation}
    \STATE $K_t \leftarrow [K_{t-1}, k_t]$
    \STATE $V_t \leftarrow [V_{t-1}, v_t]$

    \STATE \textbf{// Multi-Head Attention}
    \STATE $O_t \leftarrow \text{softmax}\left(\frac{q_t K_t^\top}{\sqrt{d}}\right)V_t$

    \STATE \textbf{// Feature Refinement}
    \STATE $s_t \leftarrow \sum_i O_t[i,:]$
    \STATE $x_{t+1} \leftarrow x_t \odot \text{Sigmoid}(s_t)$

\ENDFOR

\RETURN $\{x_{t+1}\}_{t=1}^{T}$

\end{algorithmic}
\end{algorithm}

Finally, the global token and local tokens are concatenated to form the multi-scale token set:  
\begin{equation}
X_{\text{multi}} = [x_{\text{global}}, x_{\text{local}}] \in \mathbb{R}^{(1+N_p) \times C}.
\end{equation}  

By combining holistic and localized perspectives, \( X_{\text{multi}} \) provides a richer and more diverse token space, which enhances the effectiveness of layer attention in capturing cross-layer dependencies.

\paragraph{Multi-Scale QKV Projection}

Given the multi-scale token set \( X_{\text{multi},t} \in \mathbb{R}^{(1+N_p) \times C} \) at layer \(t\), we project it into query, key, and value spaces through learnable transformations. The query is obtained by a linear mapping applied independently to each token:  
\begin{equation}
q_t = \mathbf{W}_q(X_{\text{multi},t}), \quad q_t \in \mathbb{R}^{(1+N_p) \times d},
\end{equation}
where \(\mathbf{W}_q\) is a \(1\)D convolution or fully connected layer projecting the channel dimension from \(C\) to \(d\).  

For the key representation, we apply a similar transformation:  
\begin{equation}
k_t = \mathbf{W}_k(X_{\text{multi},t}), \quad k_t \in \mathbb{R}^{(1+N_p) \times d}.
\end{equation}

For the value representation, instead of directly mapping from \(X_{\text{multi},t}\), we first enhance the original feature map \(x_t \in \mathbb{R}^{C \times H \times W}\) using a depthwise convolution \(\mathbf{W}_v\). The enriched feature is then processed by the same multi-scale pooling strategy (global scale pooling and local scale pooling) to align its granularity with the query and key:  
\begin{equation}
v_t = \text{MultiScalePool}(\mathbf{W}_v(x_t)), \quad v_t \in \mathbb{R}^{(1+N_p) \times d}.
\end{equation}

To enable efficient long-range information interaction, we progressively accumulate key and value representations across layers. Specifically, the aggregated keys and values at layer \(t\) are constructed as  
\begin{equation}
K_t = [K_{t-1}, k_t], \qquad V_t = [V_{t-1}, v_t],
\end{equation}
where \([ \cdot, \cdot ]\) denotes concatenation along the token dimension. For the first layer, we initialize \(K_0 = k_1\) and \(V_0 = v_1\).  

This progressive accumulation strategy provides each layer with access to a growing repository of multi-scale keys and values from all preceding layers, thereby facilitating richer cross-layer interactions while avoiding the quadratic complexity incurred by recomputing all tokens at every depth.  

\paragraph{Multi-Head Attention and Feature Refinement}

With the Query \(q_t \in \mathbb{R}^{(1+N_p) \times d}\) derived from the current layer's multi-scale tokens, and the accumulated Key \(K_t \in \mathbb{R}^{L_t \times d}\) and Value \(V_t \in \mathbb{R}^{L_t \times d}\) from all preceding layers, we perform multi-head attention to integrate information across layers. Here, \(L_t\) denotes the total number of accumulated tokens up to layer \(t\), including both global and local tokens from previous layers. The attention output is computed as:
\begin{equation}
O_t = \text{softmax}\left(\frac{q_t K_t^\top}{\sqrt{d}}\right) V_t, \quad O_t \in \mathbb{R}^{(1+N_p) \times d},
\end{equation}
where \(d\) is the dimensionality of each token. This allows each token in the current layer to attend to all accumulated multi-scale tokens, capturing both long-range dependencies and fine-grained information across layers.

To refine the original feature map \(x_t \in \mathbb{R}^{C \times H \times W}\), the attention output is first aggregated along the token dimension to form a channel-wise descriptor:
\begin{equation}
s_t = \sum_{i=1}^{1+N_p} O_t[i, :], \quad s_t \in \mathbb{R}^{d}.
\end{equation}
The descriptor is then passed through a sigmoid function to obtain a channel-wise attention map, which is broadcast to match the spatial dimensions of \(x_t\) and applied element-wise:
\begin{equation}
x_{t+1} = x_t \odot \text{Sigmoid}(s_t),
\end{equation}
where \(\odot\) denotes element-wise multiplication.  

This mechanism ensures that the refined feature \(x_{t+1}\) integrates information from both global and local scales across all preceding layers, producing a richer representation for subsequent processing.

\section{Experiments}

To evaluate the effectiveness of \ours for Oracle Bone Inscription (OBI) recognition, we conduct experiments on four widely used OBI benchmarks: Oracle-MNIST~\citep{wang2022oracle}, HUST-OBS~\citep{wang2024open}, OBC306~\citep{huang2019obc306}, and EVOBC~\citep{guan2024open}. A comparison of these datasets is summarized in Table~\ref{tab:dataset_comparison}. The number of character classes ranges from 10 to 13,714, covering both small-scale balanced datasets and large-scale, highly imbalanced benchmarks. Such diversity enables comprehensive evaluation of our method under varying levels of task complexity and data distribution characteristics.

\begin{table}[h]
  \centering
    \caption{Widely used datasets for Oracle Bone Inscriptions recognition.}
    \begin{tabular}{cccc}
    \hline

    \hline
    Dataset & Year & Samples & Classes   \bigstrut\\
    \hline
    OBC306 & 2019 & 300,000 & 306 \bigstrut[t]\\
    Oracle-MMNIST & 2022  & 30,222 & 10  \\
    HUST-OBS & 2024 & 140,053 & 1,588  \\
    EVOBC & 2024 & 229,170 & 13,714  \bigstrut[b]\\
    \hline

    \hline
    \end{tabular}  \label{tab:dataset_comparison}
\end{table}

\subsection{Evaluation on Oracle-MNIST}\label{experiment_oracle_mnist}

\paragraph{Dataset.} Oracle-MNIST  consists of 30,222 grayscale images of ancient Chinese characters across 10 categories, each with a resolution of 28 $\times$ 28 pixels. While it follows the MNIST format, Oracle-MNIST poses greater challenges due to severe noise caused by aging and the variability in writing styles, making it a suitable benchmark for realistic classification tasks.

\begin{table*}[!h]
  \centering
  \caption{Evaluation on Oracle-MNIST and OBC306 dataset across various methods using ResNet-20 and ResNet-56 as backbones.}

    \setlength{\tabcolsep}{18pt}
    \begin{tabular}{l|cccc}
    \hline
    
    \hline
      Datasets & \multicolumn{2}{c}{O-MNIST} & \multicolumn{2}{c}{OBC306} \\ 
    \hline
    Methods & Parameter (M) & Top-1 acc (\%) & Parameter (M) & Top-1 acc (\%) \\
    \hline
    \textbf{ResNet-20} & 0.22 & 95.09 $\pm$ 0.28 & 0.30 & 87.73 $\pm$ 0.21 \\
    SENet &    0.24  & 95.44 $\pm$ 0.25 & 0.32 & 89.42 $\pm$ 0.23 \\
    CBAM  &   0.24    &  95.30 $\pm$ 0.57 & 0.32 & 90.80 $\pm$ 0.32 \\
    ECANet &  0.22     & 95.57 $\pm$ 0.16 & 0.30 & 90.71 $\pm$ 0.37 \\
    DIANet &   0.44   & 95.53 $\pm$ 0.22 & 0.52 & 91.01 $\pm$ 0.17 \\
    MRLA-B &   0.23  & 95.97 $\pm$ 0.09 & 0.31 & 90.08 $\pm$ 0.26 \\
    MRLA-L &   0.23    &  95.95 $\pm$ 0.15 & 0.31 & 90.50 $\pm$ 0.33 \\
    DLA-B &   0.41   & 95.81 $\pm$ 0.21 & 0.48 & 90.44 $\pm$ 0.25 \\
    DLA-L &    0.41   & 95.87 $\pm$ 0.25 & 0.49 & 90.40 $\pm$ 0.33 \\ 
    \ours (Ours)  &   0.24    & \textbf{96.05 $\pm$ 0.15} & 0.31 & \textbf{91.29 $\pm$ 0.25} \\
    \hline

    \textbf{ResNet-56} &  0.59     &  95.63 $\pm$ 0.29 & 0.67 & 90.11 $\pm$ 0.16 \\
    SENet &    0.66   & 96.01 $\pm$ 0.07 & 0.73 & 90.61 $\pm$ 0.24 \\
    CBAM  &    0.66   &  95.89 $\pm$ 0.16 & 0.74 & 90.73 $\pm$ 0.28 \\
    ECANet &    0.59   & 96.00 $\pm$ 0.08 & 0.67 & 90.78 $\pm$ 0.43 \\
    DIANet &    0.81   &  96.07 $\pm$ 0.04 & 0.89 & 90.72 $\pm$ 0.24 \\
    MRLA-B &     0.62   & 94.28 $\pm$ 0.48 & 0.70 & 90.56 $\pm$ 0.34 \\
    MRLA-L &    0.62   & 95.94 $\pm$ 0.18 & 0.70 & 91.00 $\pm$ 0.21 \\
    DLA-B &     0.80  & 94.92 $\pm$ 0.54 & 0.88 & 91.12 $\pm$ 0.23 \\ 
    DLA-L &    0.80   & 96.05 $\pm$ 0.08 & 0.88 & 91.11 $\pm$ 0.32 \\
    \ours (Ours)  &    0.64   & \textbf{96.30 $\pm$ 0.11} & 0.71 & \textbf{91.77 $\pm$ 0.23} \\
    
    \hline
    
    \hline
    
    \end{tabular}
\label{tab:omnist-obc}%
\end{table*}

\paragraph{Experimental Setting.} The original grayscale images were converted to 3-channel color images and resized to 32 $\times$ 32 pixels to ensure compatibility with standard ResNet architectures. We evaluated three ResNet backbones, ResNet-20, ResNet-56, and ResNet-110, alongside several attention mechanism variants for comparison. Training was performed with a batch size of 128 over 50 epochs. Optimization employed the SGD optimizer with a momentum of 0.9 and weight decay of $1 \times 10^{-4}$. The initial learning rate was set to 0.1 and decayed by a factor of 0.1 at epochs 25 and 35 following a MultiStepLR schedule. To account for hyperparameter variability, all experiments were repeated five times, and results are reported as $mean \pm std$, where $mean$ is the average performance across runs and $std$ denotes the standard deviation. All experiments were conducted on a single NVIDIA RTX 4090 GPU.


\paragraph{Experimental Results.}
As shown in Table~\ref{tab:omnist-obc}, the proposed \ours consistently outperforms all baseline methods across different backbone networks and depths. When using ResNet-20 as the backbone, \ours achieves an accuracy of 96.05\%, surpassing the vanilla ResNet-20 by 0.96\%, while introducing only 0.02M additional parameters. Compared to other channel attention methods, such as SENet and CBAM, \ours outperforms them by 0.61\% and 0.75\%, respectively. As the network depth increases, the performance of \ours further improves, reaching accuracies of 96.30\% and 96.45\% on ResNet-56 and ResNet-110, respectively.


    
    

\begin{table*}[!t]
  \centering
    \caption{Evaluation on HUST-OBS dataset across various methods using ResNets as backbones.}
    \setlength{\tabcolsep}{18pt}
    \begin{tabular}{l|cccc}
    \hline
    
    \hline
      Datasets & \multicolumn{2}{c}{HUST-OBS} & \multicolumn{2}{c}{EVOBC} \\ 
    \hline
    Methods & Parameter (M) & Top-1 acc (\%) & Parameter (M) & Top-1 acc (\%) \\
        \hline
    \textbf{ResNet-50} &    26.78   & 93.74  & 119.29 & 49.08 \bigstrut[t]\\
    SENet &   29.30    & 93.84 & 126.53 & 48.46 \\
    CBAM  &   29.31    & 93.99 & 126.55 & 47.79  \\
    ECANet &   26.78    & 93.95 & 119.30 & 49.19 \\
    DIANet &   40.74    & 93.83 & 164.73 & 48.47 \\
    MRLA-B &    26.95   & 92.94 & 119.55 & 46.52  \\
    MRLA-L &    26.96   & 93.97 & 119.55 & 47.30  \\
    DLA-L  &   29.81    & 93.91 & 128.71 & 47.58 \\
    \ours (Ours)  &   26.96    & \textbf{94.12} & 119.56 & \textbf{49.45} \bigstrut[b]\\
    \hline
    \textbf{ResNet-101} &  45.80     & 93.83 & 138.28 & 48.68 \bigstrut[t]\\
    SENet &   50.54    & 93.87 & 147.75 & 47.35 \\
    CBAM  &   50.58    & 93.92 & 147.79 & 47.89 \\
    ECANet &    45.80   & 94.16 & 138.28 & 48.80 \\
    DIANet &   59.76    & 93.68 & 183.72 & 47.27 \\
    MRLA-L &   46.19   & 93.92 & 138.75 & 48.17  \\
    DLA-L  &  49.03     & 93.81 & 147.91 & 48.71 \\
    \ours (Ours)  & 46.19      & \textbf{94.34} & 138.76 & \textbf{49.30} \bigstrut[b]\\
    \hline
    
    \hline
    \end{tabular}
\label{tab:hust_evobc_results}%
\end{table*}%

\subsection{Evaluation on HUST-OBS}
\paragraph{Dataset.} HUST-OBS provides 140,053 images, including deciphered and undeciphered characters from 1,588 and 9,411 classes, respectively. Here, we focus on the deciphered characters with 1588 classes.

\paragraph{Experimental Setting.} We followed the official HUST-OBS implementation, adopting the same data augmentation methods. The batch size is set to 128, with a weight decay of 5e-4. The initial learning rate is set to 0.015, and it is gradually decreased using a cosine annealing schedule over 400 epochs. We evaluated the performance using ResNet-50 and ResNet-101 backbones, and compared them against multiple other models. All experiments were conducted on a single NVIDIA RTX 4090 GPU.

\paragraph{Experimental Results.}
As shown in Table \ref{tab:hust_evobc_results}, \ours achieves strong performance on the HUST-OBS dataset. With ResNet-50 as the backbone, \ours attains an accuracy of 94.12\%, surpassing the vanilla ResNet-50 accuracy of 93.74\% while introducing only 0.18M additional parameters. In comparison, the layer attention method MRLA-B underperforms the vanilla ResNet, achieving 92.94\%, indicating that relying solely on the global token to capture previous information is insufficient. By contrast, \ours improves performance by 1.18\% over MRLA-B while adding only 0.01M parameters. Similar trends are observed when using the ResNet-101 backbone, further demonstrating the effectiveness of \ours.

\subsection{Evaluation on OBC306}
\paragraph{Dataset.} OBC306  includes over 300,000 samples from 306 glyph classes, cropped from rubbings or images of oracle bones. It highlights the complexities of ancient script recognition and serves as a benchmark for automatic recognition algorithms.

\paragraph{Experimental Setting.}  
Similar to the experimental setup on Oracle-MNIST, the original grayscale images were converted to 3-channel color images and resized to 32$\times$32 pixels to ensure compatibility with standard ResNet architectures. We employed ResNet-20, ResNet-56, and ResNet-110 as backbone models and compared their performance with various methods. Given the large size of the OBC306 dataset, the batch size was set to 512, and the training process spanned 180 epochs. Optimization was performed using the SGD optimizer with a momentum of 0.9 and a weight decay of 1e-4. The initial learning rate was set to 0.20 and adjusted according to a MultiStepLR schedule, decreasing by a factor of 0.1 at epochs 100 and 150. All experiments were conducted on a single NVIDIA RTX 4090 GPU.

\begin{figure*}[htbp]
    \centering
    \begin{minipage}[b]{0.48\linewidth}
        \centering
        \includegraphics[width=\linewidth]{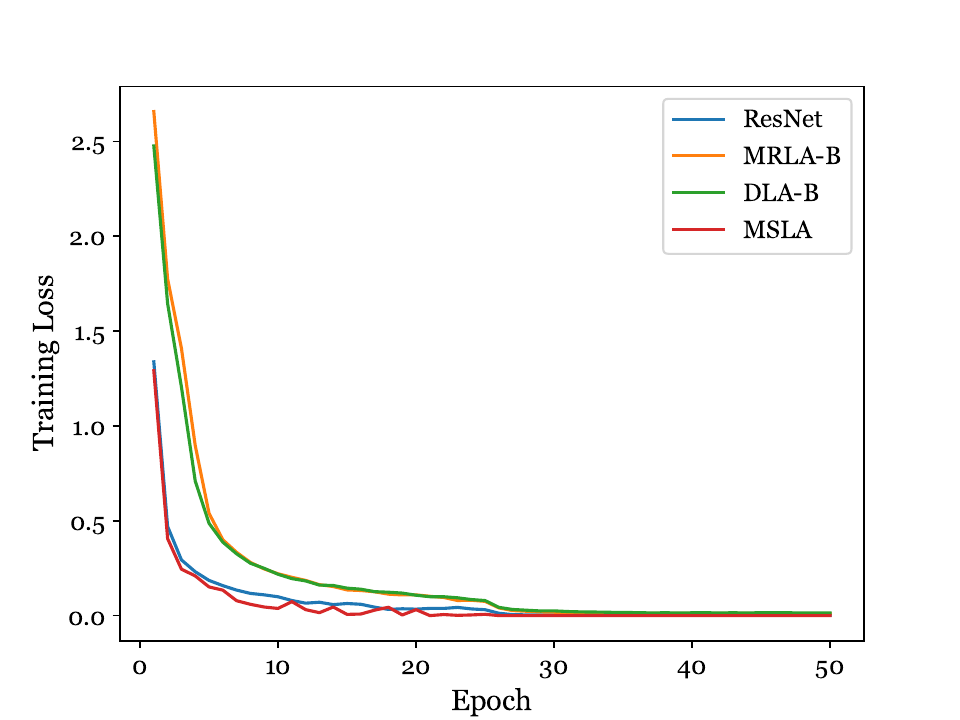}
        \subcaption{}
        \label{fig:training_loss}
    \end{minipage}
    \begin{minipage}[b]{0.48\linewidth}
        \centering
        \includegraphics[width=\linewidth]{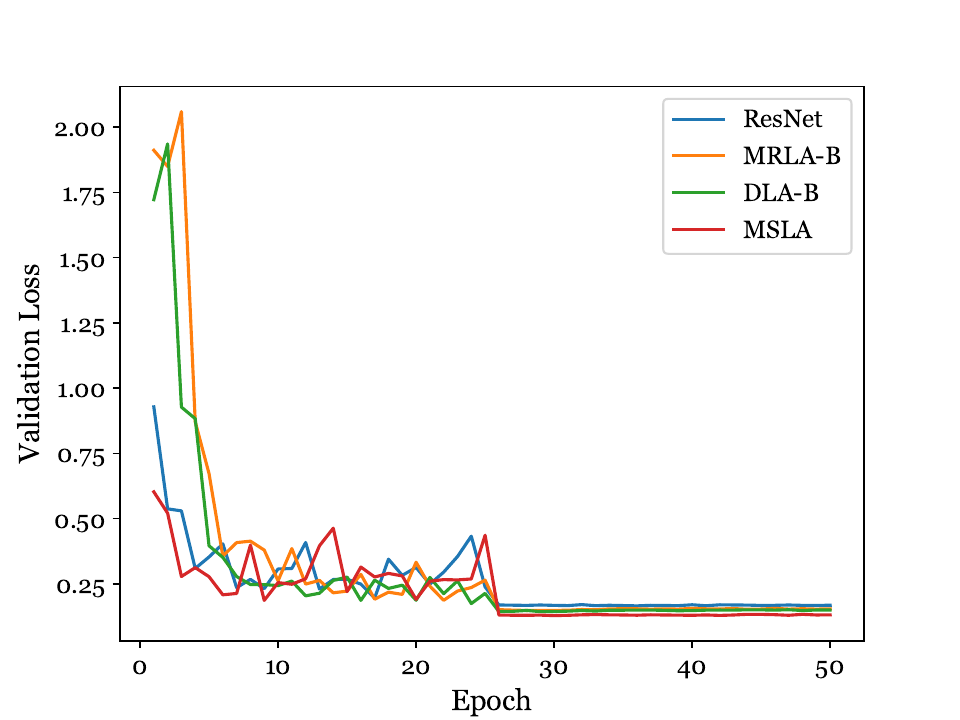}
        \subcaption{}
        \label{fig:valdation_loss}
    \end{minipage}
    \caption{Training and validation loss curves on the Oracle-MNIST dataset for various layer attention methods with ResNet-20 as the backbone.}

    \label{fig:loss_curve}
\end{figure*}

\paragraph{Experimental Results.}
As shown in Table~\ref{tab:omnist-obc}, we compare our proposed \ours with several well-known baselines. Using ResNet-20 as the backbone, \ours achieves an accuracy of 91.29\%, outperforming the vanilla ResNet-20 by 3.56\%, while introducing only 0.01M additional parameters. Surprisingly, \ours on the ResNet-20 backbone also surpasses the vanilla ResNet-56 by 1.18\%, with 0.36M fewer parameters, demonstrating the efficiency and effectiveness of our approach. Compared to channel attention mechanisms, \ours outperforms SENet, CBAM, and ECANet by 1.87\%, 0.49\%, and 0.58\%, respectively. Moreover, when compared to layer interaction methods, \ours surpasses DIANet by 0.28\% while using 0.21M fewer parameters. In addition, \ours achieves higher accuracy than MRLA-L and DLA-L, outperforming them by 0.79\% and 0.89\%, respectively.

\subsection{Evaluation on EVOBC}
\paragraph{Dataset.} EVOBC serves as one of the most challenging datasets for large-scale OBI recognition, comprising 229,170 images spanning 13,714 character categories across six historical stages. The dataset’s large scale, fine-grained categorization, and inclusion of both deciphered and undeciphered forms make it a comprehensive and demanding benchmark for evaluating ancient script recognition and reconstruction methods.

\paragraph{Experimental Setting.}
All experimental settings are the same as those used for HUST-OBS, except that the total number of training epochs is set to 200 and the initial learning rate is set to 0.1.

\paragraph{Experimental Results.}
As shown in Table~\ref{tab:hust_evobc_results}, \ours also achieves strong performance on the EVOBC dataset. Overall, all methods obtain only around 40–50\% accuracy, highlighting the difficulty and challenge of this dataset. Under these conditions, \ours still demonstrates superior performance. Using ResNet-50 as the backbone, \ours attains 49.45\% accuracy, whereas other layer interaction methods perform worse than the vanilla ResNet-50. Specifically, MRLA-B achieves only 46.52\%, compared to 49.30\% for \ours. With ResNet-101 as the backbone, overall performance decreases compared to ResNet-50; for instance, vanilla ResNet-101 performs worse than ResNet-50. Nevertheless, \ours consistently achieves the best performance among all methods under ResNet-101. These results indicate that, even on challenging datasets, \ours effectively leverages both global and local information to maintain strong recognition performance.





\subsection{Comparison with State-of-the-Art Methods}

Beyond the controlled comparisons under different attention mechanisms with unified ResNet backbones, we further compare \ours with previously published methods reported in the literature. Specifically, we collect publicly available results of both traditional machine learning approaches (e.g., SVM-based methods) and representative deep learning models evaluated on the same benchmarks.

We conduct comparisons on the Oracle-MNIST and OBC306 datasets. For fairness, we directly cite the reported Top-1 accuracy from the original papers without re-training their models. In particular, for Oracle-MNIST, all baseline results are cited from~\cite{wang2022oracle}, where multiple classical and neural methods were evaluated under different settings. We report the best-performing configuration for each method along with its corresponding parameters, as shown in Table~\ref{tab:extra_oracle}.

As observed in Table~\ref{tab:extra_oracle}, traditional machine learning models (e.g., SVC, Random Forest, Gradient Boosting) achieve moderate performance, while CNN-based methods significantly improve recognition accuracy. In comparison, \ours-20 and \ours-56 further outperform these methods, demonstrating the effectiveness of our attention-enhanced architecture.

For OBC306, we compare \ours with representative long-tailed learning methods and recent generative or enhancement-based approaches. All results are directly cited from~\citep{li2023diff,li2023towards}. As shown in Table~\ref{tab:extra_obc306}, \ours consistently achieves superior Top-1 accuracy compared with prior methods, including long-tailed optimization techniques (e.g., LWS~\citep{kang2019decoupling}, Logit Adjustment~\citep{menon2020long}, ResLT~\citep{cui2022reslt}) and generative-based approaches (e.g., AGTGAN~\citep{huang2022agtgan}, ControlNet~\citep{zhang2023adding}). This indicates that our method maintains strong performance even under the challenging distribution imbalance of OBC306.

\begin{table*}[]
    \centering
    \caption{Comparison with traditional machine learning models and classical deep learning baselines on the Oracle-MNIST dataset. All baseline results are directly cited from~\cite{wang2022oracle}, where the best-performing configuration of each method is reported together with its parameter settings.}
    \begin{tabular}{ccc}
    \hline

    \hline
    Methods & Parameter & Top-1 acc (\%) \\
    \hline
    CNN & 2 $\times$Conv-Pool-ReLu, 2$\times$FC, Dropout & 93.80\% \\
    SVC & C=10, kernel=rbf & 75.50\% \\
    MLPClassifier & activation=relu, hidden$\_$layer$\_$sizes=[100] & 74.70\%  \\
    GradientBoostingClassifier & n$\_$estimators=100, loss=deviance, max$\_$depth=10 & 72.50\% \\
    RandomForestClassifier & n$\_$estimators=100, criterion=gini, max$\_$depth=100 & 65.50\% \\
    KNeighborsClassifier & weights=distance, n$\_$neighbors=9, p=2 & 62.70\% \\
    LogisticRegression & C=10, multi$\_$class=ovr, penalty=l2 & 59.80\% \\
    LinearSVC & loss=hinge, C=1, multi$\_$class=crammer$\_$singer, penalty=l2 & 58.10\% \\
    SGDClassifier& loss=log, penalty=l1 & 56.70\% \\
    \ours-20 & - & \textbf{96.05\%} \\
    \ours-56 & - & \textbf{96.30\%} \\

    \hline

    \hline
    \end{tabular}

    \label{tab:extra_oracle}
\end{table*}

\begin{table}[]
    \centering
    \caption{Comparison with state-of-the-art methods on the OBC306 dataset.}
    \begin{tabular}{cc}

    \hline 

    \hline

    Methods & Top-1 acc (\%) \\
    \hline
    LWS~\citep{kang2019decoupling} & 84.52\% \\
    Logit adjustment post-hoc~\citep{menon2020long}& 84.42\% \\
    Logit adjustment loss~\citep{menon2020long}  & 84.63\% \\
    PaCO~\citep{cui2021parametric} & 82.28\% \\
    ResLT~\citep{cui2022reslt} & 84.05\% \\
    SADE~\citep{zhang2022self} & 86.56\% \\
     Li et al~\citep{li2023towards} & 86.54\% \\
     STSN~\cite{wang2022unsupervised} &  85.07\% \\
     AGTGAN~\cite{huang2022agtgan} & 85.99\% \\
     ControlNet~\cite{zhang2023adding} & 80.05\% \\
     ~\ours-20 & \textbf{91.29\%} \\
     ~\ours-56& \textbf{91.77\%} \\

     \hline 

     \hline

    \end{tabular}

    \label{tab:extra_obc306}
\end{table}

\section{Ablation Study}

\subsection{Evaluation on Different Patch Size $P$}
The parameter $P$ is introduced to control the granularity of multi-scale token extraction. A larger $P$ indicates that richer multi-scale information is extracted from a single feature, whereas a smaller $P$ corresponds to a coarser representation. To further investigate the impact of $P$, we conduct experiments on the OBC306 dataset with different patch sizes, where $P$ is varied from 1 to 5. Notably, when $P=1$, the proposed \ours degenerates to MRLA-B.

\begin{figure*}[htbp]
    \centering
    \begin{minipage}[b]{0.48\linewidth}
        \centering
        \includegraphics[width=\linewidth]{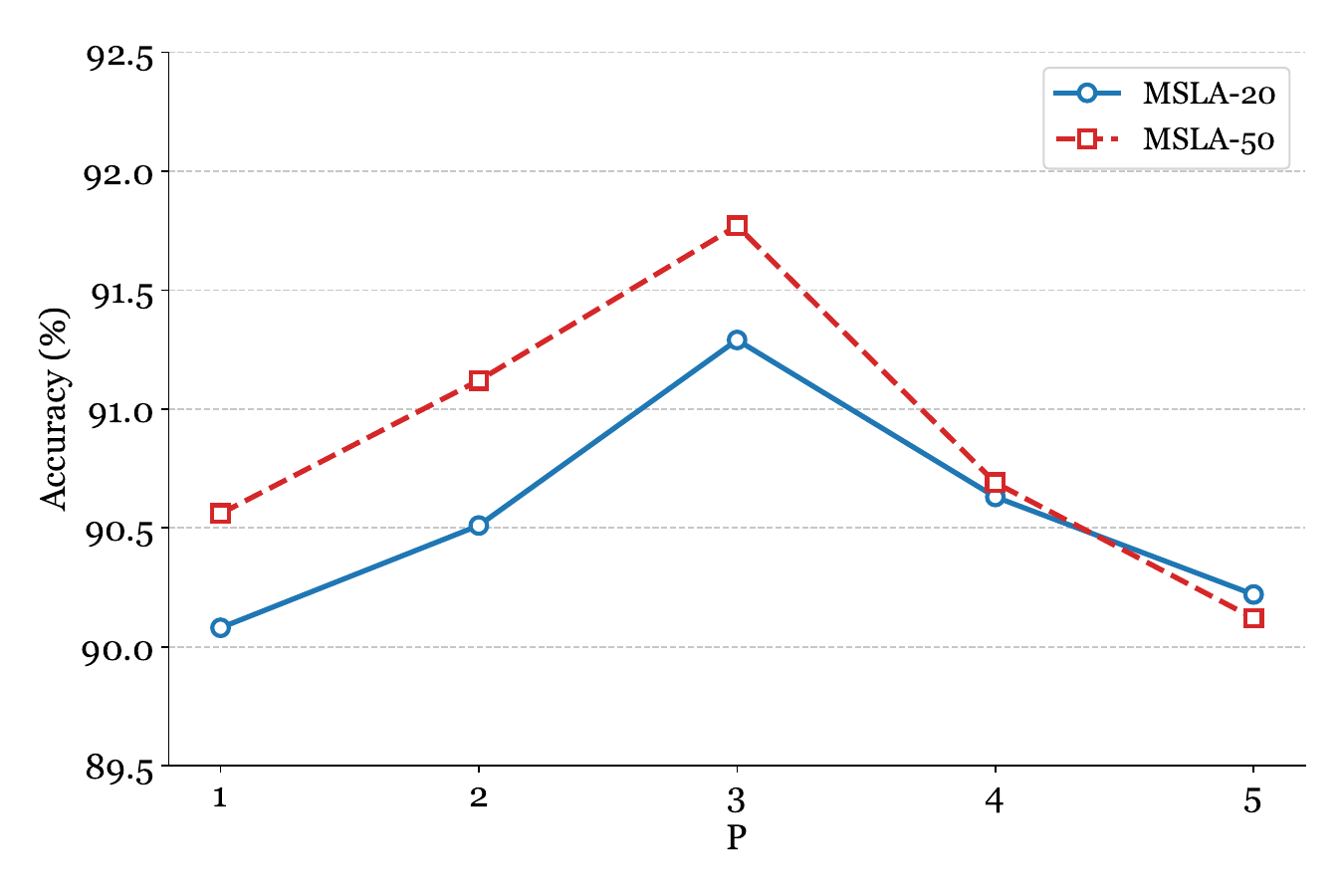}
        \subcaption{}
        \label{fig:ablation1}
    \end{minipage}
    \begin{minipage}[b]{0.48\linewidth}
        \centering
        \includegraphics[width=\linewidth]{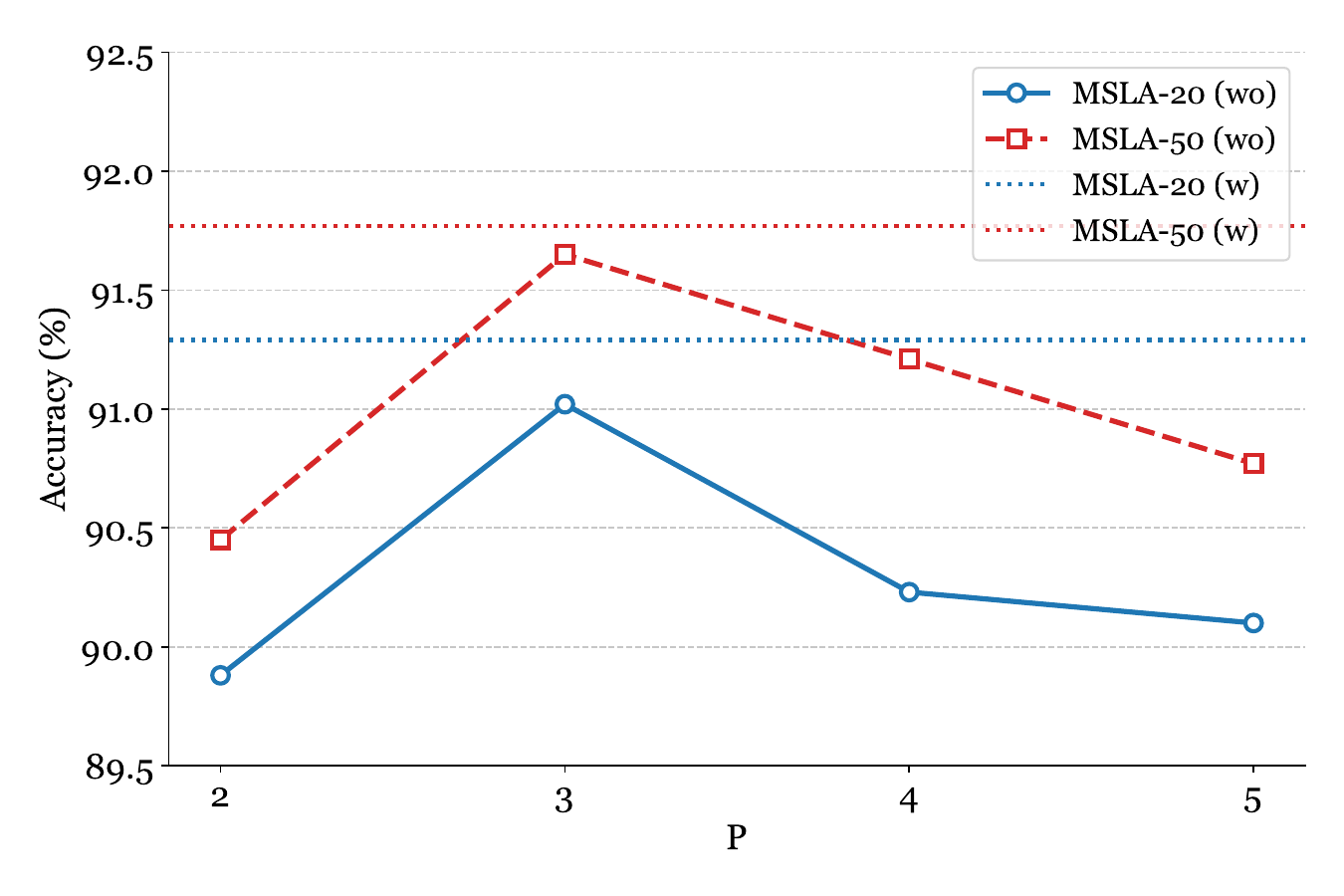}
        \subcaption{}
        \label{fig:ablation2}
    \end{minipage}
    \caption{Ablation studies: (a) Evaluation on different patch size $P$ and (b) Evaluation on local tokens.}
    \label{fig:ablation_all}
\end{figure*}

As shown in Figure~\ref{fig:ablation1}, \ours achieves the best performance when $P=3$. As $P$ increases from 1 to 3, the performance consistently improves, indicating that incorporating richer multi-scale information effectively enhances the representation for OBI recognition. However, when $P$ further increases from 3 to 5, the performance begins to degrade. This suggests that excessive scale information may introduce redundancy and noise, which in turn hampers effective feature aggregation and negatively affects recognition performance.

\subsection{Evaluation on Different Input Sizes}
Due to limited computational resources, we resize the input images to $32\times32$ by default. To further verify the generalization capability of our proposed method, we conduct experiments with different input resolutions, including $28\times28$, $32\times32$, $64\times64$, $128\times128$, and $256\times256$. All experiments are performed on the Oracle-MNIST dataset.

\begin{table}[t]
  \centering
    \caption{The comparison of different input sizes across \ours on O-MNIST dataset.}
    \begin{tabular}{lccccc}
    \hline

    \hline
    Models  & 28 & 32 & 64 & 128 & 256 \bigstrut\\
    \hline
    \ours-20   & 96.00 & 96.05 & 96.23 & 96.26 & 96.31  \bigstrut[t]\\
    \ours-56 & 96.19 & 96.30 & 96.35 & 96.66 & 96.72  \bigstrut[b]\\
    \hline

    \hline
    \end{tabular}  \label{tab:input_size}
\end{table}

As shown in Table~\ref{tab:input_size}, the performance of both \ours-20 and \ours-56 consistently improves as the input size increases. These results indicate that \ours not only achieves strong performance under the default setting but also maintains excellent effectiveness across different input sizes, demonstrating its robustness and generalizability.

\subsection{Evaluation on Local Tokens}
As discussed earlier, when $P=1$, \ours degenerates to MRLA-B, where only the global token is used. To further investigate the contribution of local tokens, we conduct additional experiments on the OBC306 dataset by removing the global token and retaining only local tokens. Specifically, we evaluate different settings of $P \in \{2,3,4,5\}$ under this configuration.

As shown in Figure~\ref{fig:ablation2}, even without the global token, the performance still varies with $P$, reaching the best result when $P=3$. However, removing the global token leads to an overall performance drop, demonstrating the importance of the global token in capturing holistic context information that complements local tokens.

\section{Conclusion}
In this paper, we study the challenging problem of oracle bone inscriptions (OBIs) recognition and analyze the limitations of existing attention mechanisms in capturing fine-grained details and complex structural variations. In particular, conventional attention and recent layer attention methods are constrained by limited token diversity in cross-layer interactions, which restricts their effectiveness in OBI recognition. To address this issue, we propose \textbf{Multi-Scale Layer Attention (\ours)}, a novel attention paradigm that explicitly models both multi-scale and cross-layer feature interactions. By enriching the token space with multi-scale representations and progressively aggregating information across layers, \ours enhances the model’s ability to capture subtle variations and structural dependencies in OBIs. Extensive experiments on large-scale OBI datasets demonstrate that \ours consistently outperforms existing attention mechanisms while maintaining high computational efficiency. These results validate the effectiveness and robustness of the proposed method for fine-grained OBI recognition and highlight its potential applicability to other complex visual recognition tasks.

\section*{Declaration of Competing Interest}
The authors declare that there is no competing or conflicting interest.


\section*{Declaration of Generative AI and AI-assisted Technologies}

During the preparation of this work, the authors used ChatGPT solely to improve language clarity and readability. All other aspects of this work, including research ideas, data collection, analysis, interpretation, and code, were conducted entirely by the authors without the use of AI tools. 

\section*{Acknowledgments}
This work was supported by the Natural Science Foundation of Sichuan Province under Grant No. 2026NSFSC1488, the Social Science Foundation of Sichuan Province under Grant No. SCJJ25ND235, and the Social Science Foundation of China under Grant No. 24FYYB049.








\printcredits

\bibliographystyle{cas-model2-names}

\bibliography{cas-refs}



\end{document}